\documentclass{article}
\usepackage{acl2000}
\usepackage{epsfig}
\usepackage{floatflt}

\title{\vspace{-65pt}{\normalsize \tt \hfill Appeared in
\em{Proceedings of the 38th ACL}, 2000}\\ \mbox{} \\ Rule Writing or
Annotation: Cost-efficient Resource Usage \\ for Base Noun Phrase
Chunking} 

\author{
	Grace Ngai \and David Yarowsky\\
	Department of Computer Science \\
	Johns Hopkins University\\
	Baltimore, MD 21218, U.S.A.\\
	Email:{\tt \{gyn,yarowsky\}@cs.jhu.edu}
	}
\date{}

\begin{document}

\maketitle

\begin{abstract}
This paper presents a comprehensive empirical comparison between two
approaches for developing a base noun phrase chunker: human rule
writing and active learning using interactive real-time human
annotation.  Several novel variations on active learning are
investigated, and underlying cost models for cross-modal machine
learning comparison are presented and explored. Results show that it
is more efficient and more successful by several measures to train a
system using active learning annotation rather than hand-crafted rule
writing at a comparable level of human labor investment.
\end{abstract}

\section{Introduction}

One of the primary problems that NLP researchers who work in new
languages or new domains encounter is a lack of available annotated
data.  Collection of data is neither easy nor cheap.  The construction
of the Penn Treebank significantly improved performance for English
systems dealing in the ``traditional'' NLP domains (eg parsing,
part-of-speech tagging, etc).  However, for a new language, a similar
investment of effort in time and money is most likely prohibitive, if
not impossible.

Faced with the costs associated with data acquisition, rationalists
may argue that it would be more cost effective to construct systems of
hand-coded rule lists that capture the linguistic characteristics of
the task at hand, rather than spending comparable effort annotating
data and expecting the same knowledge to be acquired indirectly by a
machine learning system.  The question we are trying to address then
is: for a given cost assumption, which approach would be the most
effective.

Although learning curves showing performance relative to amount of
training data are common in the machine learning literature, these are
inadequate for comparing systems with different sources of training
data or supervision.  This is especially true when a human rule-based
approach and empirical learning are evaluated relative to effort
invested.  Such a multi-factor cost analysis is long overdue.

This paper will conclude with a comprehensive cost model exposition
and analysis, and an empirical study contrasting human rule-writing
versus annotation-based learning approaches that are sensitive to
these cost models.

\section{Base Noun Phrase Chunking}

The domain in which our experiments are performed is base noun phrase
chunking.  A significant amount of work has been done in this domain
and many different methods have been applied: Church's PARTS
\shortcite{church88:PARTS} program used a Markov model; Bourigault
\shortcite{bourigault92:basenp} used heuristics along with a grammar;
Voutilainen's NPTool \shortcite{voutilainen:NPTool} used a lexicon
combined with a constraint grammar; Juteson and Katz
\shortcite{juteson95:basenp} used repeated phrases; Veenstra
\shortcite{veenstra98:basenp}, Argamon, Dagan \& Krymolowski
\shortcite{argamon98:basenp}, Daelemans, van den Bosch \& Zavrel 
\shortcite{daelemans99:exceptions} and Tjong Kim Sang \& Veenstra
\shortcite{tjong99:basenp} used memory-based systems; Ramshaw \& Marcus 
\shortcite{ramshaw99:basenp} and Cardie \& Pierce
\shortcite{cardie98:basenp} used rule-based systems, Munoz et al.
\shortcite{roth99:basenp} used a Winnow-based system, and the XTAG
Research Group\shortcite{xtag98:basenp} used a tree-adjoining grammar.

Of all the systems, Ramshaw \& Marcus' transformation rule-based
system had the best published performance (f-measure 92.0) for several
years, and is regarded as the de facto standard for the domain.
Although several systems have recently achieved slightly higher
published results (Munoz et al.: 92.8, Tjong Kim Sang \& Veenstra:
92.37, XTAG Research Group: 92.4), their algorithms are significantly
more costly, or not feasible, to implement in an active learning
framework.  To facilitate contrastive studies, we have evaluated our
active learning and cost model comparisons using Ramshaw \& Marcus'
system as the reference algorithm in these experiments.

\section{Active Learning from Annotation}

Supervised statistical machine learning systems have traditionally
required large amounts of annotated data from which to extract
linguistic properties of the task at hand.  However, not all data is
created equal.  A random distribution of annotated data contains much
redundant information.  By intelligently choosing the training
examples which get passed to the learner, it is possible to provide
the necessary amount of information with less data.

Active learning attempts to perform this intelligent sampling of data
to reduce annotation costs without damaging performance.  In general,
these methods calculate the usefulness of an example by first having
the learner classify it, and then seeing how uncertain that
classification was.  The idea is that the more uncertain the example,
the less well modeled this situation is, and therefore, the more
useful it would be to have this example annotated.

\subsection{Prior Work in Active Learning}

Seung, Opper and Sompolinsky \shortcite{seung92:query_by_committee}
and Freund et al. \shortcite{freund97:query_by_committee} proposed a 
theoretical {\em query-by-committee} approach.  Such an approach uses
multiple models (or a {\em committee}) to evaluate the data, and
candidates for annotation (or {\em queries}) are drawn from the pool
of examples in which the models disagree.  Furthermore, Freund et
al. prove that, under some situations, the generalization error
decreases exponentially with the number of queries.

On the experimental side, active learning has been applied to several
different problems.  Lewis \& Gale
\shortcite{lewis94:active_learning}, Lewis \& Catlett
\shortcite{lewis94:heterogeneous} and Liere \& Tadepalli
\shortcite{liere97:active_learning_textcat} all applied it to
text categorization; Engelson \& Dagan
\shortcite{engelson96:active_learning_POS} applied it to
part-of-speech tagging.

Each approach has its own way of determining uncertainty in examples.
Lewis \& Gale used a probabilistic classifier and picked the examples
$e$ whose class-conditional {\em a posteriori} probability $P(C \vert
e)$ is closest to 0.5 (for a 2-class problem).  Engelson \& Dagan
implemented a committee of learners, and used vote entropy to pick
examples which had the highest disagreement among the learners.  In
addition, Engelson \& Dagan also investigate several different
selection techniques in depth.

\begin{figure*}[htb]
\begin{center}
%\begin{tabular}{cc}
\epsfig{file=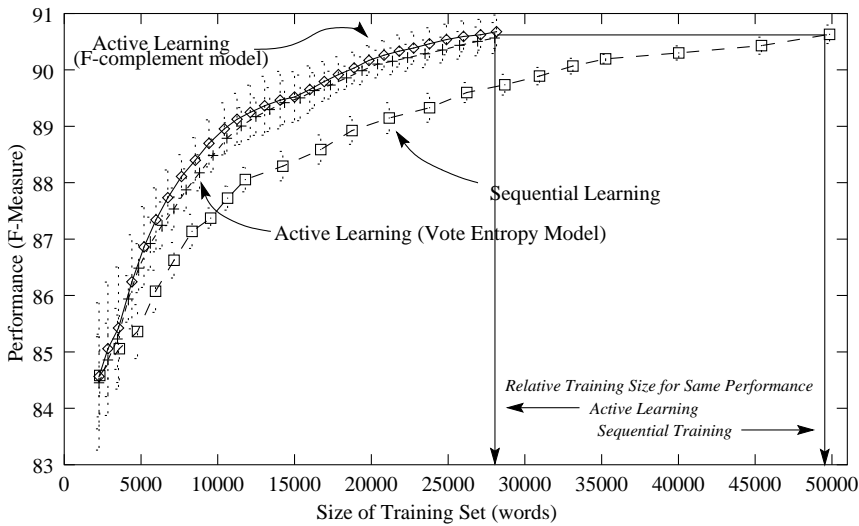, width=4.5in}% &
%\epsfig{file=al_loglog, width=3in}\\
%\end{tabular}
\caption{ \label{figure:al_performance} Performance vs. training
set size: active learning and sequential annotation on Treebank data}
\end{center}
\end{figure*}

\subsection{New Applications and Algorithmic Extensions in Active
Learning} 

To our knowledge, this paper constitutes the first work to apply
active learning to base noun phrase chunking, or to apply active
learning to a transformation-learning paradigm
\cite{brill95:transform_nlp} for any application.
Since a transformation-based learner does not give a probabilistic
output, we are not able to use Lewis \& Gale's method for determining
uncertainty.  Our experimental framework thus uses the query by
committee paradigm with batch selection:

\begin{enumerate}
\item Given a corpus $C$, arbitrarily pick $t$ sentences for
annotation.  
\item Have these $t$ sentences hand-annotated, delete them from
$C$ and put them into a training set, $T$.
\item \label{algorithm:al_start} Divide $T$ into $m$ non-identical,
but not necessarily non-overlapping, subsets. 
\item Use each subset as the training set for a model.
\item Evaluate each model on the remaining sentences in $C$
\item Using a measure of disagreement $D$, pick the $x$ sentences in
$C$ with the highest $D$ for annotation.
\item Delete the $x$ sentences from $C$, have them annotated, and add
them to $T$.
\item Repeat from \ref{algorithm:al_start}.
\end{enumerate}

In our experiments, the initial corpus $C$ that we used consisted of
sections 15-18 of the Wall Street Journal Treebank
\cite{marcus93:penn_treebank}, which is also the training set used by
Ramshaw \& Marcus \shortcite{ramshaw99:basenp}.  The initial $t$
sentences were the first 100 sentences of the training corpus, and
$x=50$ sentences were picked at each iteration.  Sets of 50 sentences
were selected because it takes approximately 15-30 minutes for humans
to annotate them, a reasonable amount of work and time for the
annotator to spend before taking a break while the machine selects the
next set. The parameter $m$, which denotes the number of models to
train, was set at 3, which could be expected to give us reasonable
labelling variation over the samples, but also would not cause the
processing phase to take a long time.

To divide the corpus into the different subsets in Step
\ref{algorithm:al_start}, we tried using two approaches: bagging and
n-fold partitioning.  In bagging, we randomly pick (with replacement)
$\frac{2}{3}$ of the total number of sentences in $C$ to assign to
each subset.  With n-fold partitioning, we partitioned the data into 3
discrete partitions, and each model was then trained on 2 of the 3
partitions.  We found no significant difference between the two
methods. 

\subsubsection{Models of Disagreement for the Selection of New Data}

The standard method for measuring disagreement for sample selection in 
active learning algorithms that use the query by committee is Engelson 
\& Dagan's vote entropy measure.  Given a tagged example
$e$\footnotemark[1], the disagreement $D$ for $e$ is\footnotemark[2]:

\footnotetext[1]{Vote entropy calculates the disagreement on a per
tagged unit basis.  In domains such as part-of-speech tagging or base
noun phrase chunking, each tagged unit is a word.  We prefer to select
entire sentences as candidates for annotation.  In situations like
these, the disagreement over the entire sentence is simply the mean 
disagreement over the words in the sentence.}

\footnotetext[2]{Dividing by log $k$ normalizes for the number of models}

\begin{eqnarray*}
D &=& -\frac{1}{\mbox{log }
k}\sum\limits_{c}\frac{V(c,e)}{k}\mbox{log}\frac{V(c,e)}{k}\\
\mbox{where}\\
k &=& \mbox{Number of models in the committee.}\\
V(c,e) &=& \mbox{Number of models assigning $c$ to $e$}
\end{eqnarray*}

However, here we propose a novel disagreement measure that is both
more applicable and achieves slightly improved performance.  We base
our measure on the f-measure metric, which is defined as:
\begin{eqnarray*}
F_\beta &=&
\frac{(\beta^2+1)\times\mbox{Precision}\times\mbox{Recall}}{\beta^2\times\mbox{Precision}+\mbox{Recall}}\\
\mbox{where}\\
\mbox{Precision} &=& \frac{\mbox{\# of correct
proposed labellings}}{\mbox{\# of proposed labellings}}\\
\mbox{Recall} &=& \frac{\mbox{\# of correct proposed
labellings}}{\mbox{\# of correct labellings}}
\end{eqnarray*}
The variable $\beta$ allows precision and recall to be weighed
differently.  In all our experiments, $\beta$ is set to 1, giving an
equal weight to both precision and recall.  

For our disagreement measure $D$, we use the {\em f-complement}, which 
is calculated as:\\
%\begin{eqnarray*}
\[
D = \frac{1}{2} \sum\limits_{M_i,M_j \in \mathcal{K}} 
\left( 1-F_1(M_i(e), M_j(e)) \right) 
\]
where
$\mathcal{K}$ is the committee of models, 
$M_i, M_j$ are individual models in $\mathcal{K}$, and
$F_1(M_i(e),M_j(e))$ is F$_1$ of $M_i$'s
labelling of $e$ relative to $M_j$'s evaluation of
$e$.\footnotemark[3] 
%\end{eqnarray*}

\footnotetext[3]{$\beta=1$ makes the F-measure symmetrical.} 

Figure \ref{figure:al_performance} shows the test set performance
against the number of words in the training corpus for sequential
annotation and active learning, using vote entropy and f-complement as
the measures of disagreement.  As can be seen from the graphs,
f-complement gives a small empirical boost in performance.  More
importantly, f-complement can be used in applications where
implementation of vote entropy is difficult, for example, parsing. The
comparison between systems trained on annotated sentences selected by
active learning and annotated sentences selected sequentially shows
that active learning reduces the amount of data needed to reach a
given level of performance by approximately a factor of two.  

%Figure
%\ref{figure:al_log_performance} illustrates that performance growth
%continues log-linearly, and gives an estimate of how much data would
%be needed to achieve a certain desired level of performance.
%\begin{figure}[htb]
%\begin{center}
%%\begin{tabular}{cc}
%\hspace*{-0.3in}\epsfig{file=al_loglog, width=3.5in}% &
%%\epsfig{file=al_loglog, width=3in}\\
%%\end{tabular}
%\caption{ \label{figure:al_log_performance} Performance versus Training
%Set Size: Log-log Scale}
%\end{center}
%\end{figure}

\subsection{\label{realtime_al}Active Learning with Real Time Human
Supervision} 

Most of the published work on active learning are simulations of an
idealized situation.  One has a large annotated corpus, and the new
tags for the ``newly annotated'' sentences are simply drawn from what
was observed in the annotated corpus, as if the gold standard
annotator was producing this feedback in real time, while the test
set itself is, of course, not used for this feedback.  This is an
idealized situation, since it assumes that a true active learning
situation would have access to someone who could annotate with perfect 
consistency to the gold standard corpus annotation conventions.

Because our goal is to investigate the relative costs of rule writing
versus annotation, it is essential that we use a realistic model of
annotation.  Therefore, we decided to do a fully-fledged active
learning annotation experiment, with real time human supervision,
rather than assume the simulated feedback of actual Treebank
annotators.

We developed an annotation tool that is modeled on MITRE's Alembic
Workbench software \cite{day97:alembic}, but written in Java for
platform-independence.  To enable data storage and the active learning
sample selection to take place on the more powerful machines in our
lab rather than the user's home machine, the tool was designed with
network support so that it could communicate with our servers over the
internet.  %Figure \ref{figure:annotationTool} is a screenshot of the
%system. 

%\begin{figure*}
%\begin{center}
%\mbox{\epsfig{file=annotationTool, width=5.5in}}
%\caption{ \label{figure:annotationTool} Screenshot of Annotation Tool
%for Active Learning }
%\end{center}
%\end{figure*}

Our real-time active learning experiment subjects were seven graduate
students in computer science.  Five of them are native English
speakers, but none had any formal linguistics training.  The initial
training set $T$ is the first 100 sentences of Ramshaw \& Marcus'
training set.  To acquaint the subjects with the Treebank conventions,
they were first asked to spend some time in a feedback phase, where
they would annotate up to 50 sentences (they were allowed to stop at
any time) drawn from the initial 100 sentences in $T$.  The sentences
were annotated one at a time, and the Treebank annotation was shown to
them after every sentence.  On average, the annotators spent around 15
minutes on this feedback phase before deciding that they were
comfortable enough with the convention.

The active learning phase follows the feedback phase.  The
f-complement disagreement measure was used to select 50 sentences from
the rest of Ramshaw \& Marcus' training set and the annotator was
instructed to annotate them.  The annotated sentences were then sent
back to the server.  The system chose the next 50 sentences.  The
experiment consists of 10 iterations, during which the annotators were
allowed to make use of the original 100 sentences as a reference
corpus.  After completing all 10 iterations, they were asked to
annotate a further 100 consecutive sentences drawn randomly from the
test set.  The purpose of this final annotation was to judge how well
annotators tag sentences drawn with the true distribution from the
test corpus, as we shall see in section \ref{section:compare_results}.

On average, the annotators took 17 minutes to annotate each set of 50
sentences, ranging from 8 to 30 minutes.  The average amount of time
the server took to run the active learning algorithm and select the
next batch of sentences was approximately 3 minutes, a rest break for
the annotators.

The analysis of the results is presented in section
\ref{section:compare_results}.

\section{Learning by Rules}

In previous work, Brill \& Ngai \shortcite{ngai99:man_machine} showed
that under certain circumstances, it is possible for humans writing
rules to perform as well as a state-of-the-art machine learning system
for base noun phrase chunking.  What that study did not address,
however, was the cost of the human labor and/or machine cycles
involved to construct such a system, nor the relative cost of
obtaining the training data for the machine learning system.  This
paper will estimate and contrast these costs relative to performance.

\begin{table*}
\begin{scriptsize}
%\begin{tt}
\hspace*{-0.2in}\begin{tabular}{|@{\hspace{2pt}}c@{\hspace{2pt}}|l@{\hspace{1pt}}l|l@{\hspace{1pt}}l|l@{\hspace{1pt}}l|}
\hline
\hline
 & \multicolumn{6}{|c|}{\rm Example Task} \\
\cline{2-7}
 & \multicolumn{2}{c|}{\rm Inserting New Brackets} &
\multicolumn{2}{|c|}{\rm Splitting A Noun Phrase} &
\multicolumn{2}{|c|}{\rm Moving A Bracket}\\
\hline
\hline
 & & & & & & \\
 & {\sc{\bf TY:}} & A & {\sc{\bf TY:}} & S & {\sc{\bf TY:}} & T \\
{\rm Brill \& Ngai} & {\sc{\bf LC:}} & null & {\sc{\bf LC:}} & null & {\sc{\bf LC:}}&
$<<<$ (\{1\} w=about) \\
{\rm 1999 } & {\sc{\bf TAR:}} & (\{1\} t=DT) (*
t=JJ[RS]?) $\backslash$& {\sc{\bf TAR1:}} & (* t=$\backslash$w+) (+
t=NNP?S?) &  {\sc{\bf TAR:}} & (\{1\} t=\$) (+ t=CD) \\%$
{\rm Rule Format} & &
\multicolumn{1}{r|}{(+ t=NNP?S?)} & 
{\sc{\bf MC:}} & null & {\sc{\bf RC:}} & null \\ 
 & {\sc{\bf RC:}} &
null & {\sc{\bf TAR2:}} & (* t=JJ[RS]?) (\{1\} w=$\backslash$w+day) & & \\
& & & {\sc{\bf RC:}} & null & &\\
 & & & & & & \\
\hline
\hline
{\rm New} & & & & & &\\
{\rm Rule}&\multicolumn{2}{c|}{{\sc \{ \_DT ADJ* NOUN+ \}}} &
\multicolumn{2}{c}{\sc{[ \{ ANYWORD* NOUN+ \} \{ ADJ* TIMEDAY \} ]}} &
\multicolumn{2}{|c|}{\{ about\_ [ \_\$ NUM+ ] \}}\\  
\multicolumn{1}{|c|}{\rm Format}& & & & & &\\
\hline
\hline
 & & & & & & \\
{\rm Effect of} & \multicolumn{2}{c|}{\rm The$_{DT}$ man$_{NN}$
ran$_{VBD}$ .$_.$ }  & \multicolumn{2}{c|}{\rm ( New$_{NNP}$
York$_{NNP}$ Friday$_{NNP}$ ) } & \multicolumn{2}{c|}{\rm about$_{IN}$
( \$$_\$$ 5$_{CD}$ ) } \\ 
{\rm Rule } &
\multicolumn{2}{c|}{$\Downarrow$} &
\multicolumn{2}{c|}{$\Downarrow$} & 
\multicolumn{2}{c|}{$\Downarrow$} \\
{\rm Application } & \multicolumn{2}{c|}{\rm ( The$_{DT}$ man$_{NN}$ )
ran$_{VBD}$ .$_.$ }  & \multicolumn{2}{c|}{\rm ( New$_{NNP}$
York$_{NNP}$ ) ( Friday$_{NNP}$ ) } & \multicolumn{2}{c|}{\rm (
about$_{IN}$ \$$_\$$ 5$_{CD}$ ) } \\ 
 & & & & & & \\
\hline
\hline
\end{tabular}
%\end{tt}%$
\end{scriptsize}
\caption {\label{table:humanRules} Comparison of our current rule
format with Brill \& Ngai \shortcite{ngai99:man_machine}}
\end{table*}

To investigate the costs of a human rule-writing system, we used a
similar framework to that of Brill \& Ngai.  The system was written as
a cgi script which could be accessed across the web from a browser
such as Netscape or Internet Explorer.  Like Brill \& Ngai's 1999
approach, our rules were based on Perl regular expressions.  However,
instead of explicitly defining rule actions and having different kinds
of rules, our rules implicitly define their actions by using different
symbols to denote the placement of the base noun phrase-enclosing
parentheses prior to and after the application of the rule.  Table
\ref{table:humanRules} presents a comparison of our rule format
against that of Brill \& Ngai's.  The rules presented here may be
considered less cumbersome and more intuitive.

In a way that is similar to Brill \& Ngai's system, our rules were
translated into Perl regular expressions and evaluated on the corpus.
New rules are appended onto the end of the list and each rule applied
in order, in the paradigm of a transformation-based rule list.

\subsection{Rule-Writing Experiments}

The rule-writing experiments were conducted by a group of 17 advanced
computer science students, using the identical test set as in the
annotation experiments and the same initial 100 gold standard
sentences for both initial bracketing standards guidance and
rule-quality feedback throughout their work.

\begin{figure}[htb]
\begin{small}
\begin{verbatim}
# Grab-all rule
{ _RB::? ADJ* ANOUN* ADJ* ANOUN+ }
# (blah blah last Fri)->(blah blah) (last Fri)
{ [ ANYTHING* } { _JJ TIME_W ] }
{ [ NOT_ADJ+ } { TIME_W ] }
# about $8 (an ounce) -> (about $8 an ounce)
{ (Only|only|About|about)_::? _(\$|#)::?  \
   _CD::+ [ ANYTHING+ ] }
{ _RBR::* _(PDT|JJ)::? _(DT|PRP\$|POS) ADJ* \ 
   _RB::? VERB? [ ANYTHING+ ] }
# ( boy ) -> ( that boy )
[ (That|that)__DT { ANYTHING+ } ]
# ( about 4 1/2 )
{ (only|about)_::? (\$|#)_::? _CD::+ }
{ [ ANYTHING+ [? ANYTHING* ]? ] _-LRB-  \
   [ ANYTHING+ ]  _-RRB- [ ANYTHING+ ] }
# Pronouns are usually baseNPs
{ _DT::? _PRP }
# ``and'' usually isn't in a baseNP
{ [ _\S+::+ ] (and|&)_ [ _\S+::+ ] }
# more singleton baseNPs
{ _(DT|EX|WP|WDT) } VERB
# some numbers are singleton baseNPs
{ [ ANYTHING ] [ _CD ] }
# ( much/most ) of
{ _(DT|RB)::? (much|most)_ } _IN
\end{verbatim}
\end{small}
\caption{\label{figure:rulelist} An Example Rule List.  Lines
beginning with hash marks (\#) are comments.} 
\end{figure} 

The time that the students spent on the task varied widely, from a
minimum of 1.5 hours to a maximum of 9 hours, with an average of 5
hours.  Because we captured and saved every change the students made
to their rule list and logged every mouse click they made while doing
the experiment, it was possible for us to trace the performance of the
system as a function of time.  Figure
\ref{figure:rulelist} shows the rule list constructed by one of the
subjects.  The quantitative results of the rule-writing experiments
are presented in the next section.

%Figure \ref{figure:rules-performance} shows the performance of the
%human constructed rule lists as a function of the time spent writing
%rules for the top 6 humans.  

\section{\label{section:compare_results}Experiment Results \,---\,
Rule Writing vs. Annotation} 

%\begin{figure*}[h]
%\begin{center}
%\mbox{\epsfig{file=rules-performance,height=4in}}
%\caption{\label{figure:rules-performance}Performance versus Time
%Spent Rule Writing}
%\end{center}
%\end{figure*}

This section will analyze and compare the performance of systems
constructed with hand-built rules with systems that were trained
from data selected during real-time active learning.

\begin{figure*}[htb]
\begin{center}
\mbox{\epsfig{file=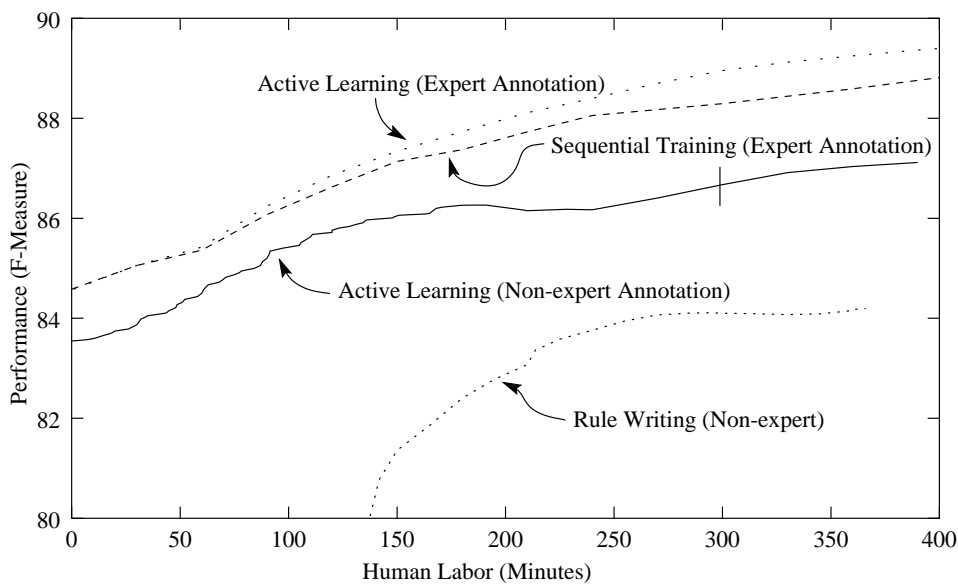, width=5in}}
\caption{\label{figure:annotation-average} Consensus Performance for
Systems constructed via Rule Writing, non-expert Annotation and
Treebank Annotation (Individual curves are in Figure
\ref{figure:annotation-ruleWriting})}
\end{center}
\end{figure*}

\begin{figure}[htb]
\begin{center}
\mbox{\epsfig{file=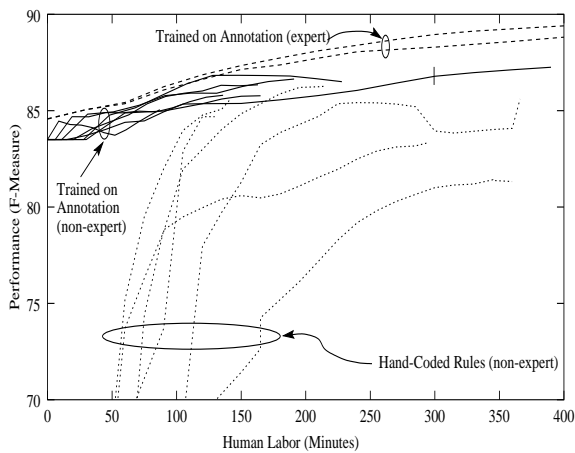, width=3in, height=2.3in}}
\caption{\label{figure:annotation-ruleWriting} Annotation versus Rule
Writing: Performance detailed by individual participant.}
\end{center}
\end{figure}

The performance of Ramshaw \& Marcus' system trained on the
annotations of each subject in the real-time active learning
experiments, and the performance achieved by the manually constructed
systems of the top 6 rule writers are shown in Figures
\ref{figure:annotation-average} and
\ref{figure:annotation-ruleWriting}, depicting the performance
achieved by each individual system.  The x-axes show the time spent
by each human subject (either annotating or writing rules) in
minutes; the y-axes show the f-measure performance achieved by the
systems built using the given level of supervision.

\subsection{Analysis of Comparative Experimental Data}

It is important to note that when comparing the curves in Figure
%\ref{figure:rules-performance}-
\ref{figure:annotation-ruleWriting},
experimental conditions across groups were kept as equal as possible,
with any known potential biases favoring the rules-writing
group. First, both groups began with the identical 100 sentence gold
standard set, for initial inspection and performance feedback
throughout the rule-writing process. The higher starting point for the
annotation-driven learning curves was due to the fact that the machine
learning algorithm could do initial training immediately on this
data. The rule-writing learners also received immediate feedback on
their first rules using this data, but were slower to incorporate this
feedback into their new rules.  The six rule-writers used for
comparative purposes were all native speakers, while the annotation
group included 2 non-native speakers.
%The annotation group included 2 non-native speakers, while all
%rule-writers evaluated were native speakers. 
Also, to further minimize the potential for any unknown biases in
sample selection in favor of annotation, the rule-writers who were
evaluated and illustrated in these graphs were the 6 strongest
performers out of the pool of 17; while all 7 annotation results are
compared. Despite this favorable treatment, rule-writing still
underperforms annotation-based learning with statistical significance
of $P<0.02$ for 100 minutes of investment, and with significance of
$P<0.05$ for times up to at least 2.5 hours.  The high variance in the
rule-writer pool complicates a finding of significance beyond this
point, but at all quantities of human labor invested, mean
annotation-based F-measure outperformed rule-writing and these trends
appear to extrapolate.
 
\subsection{Analysis of Human Performance on the Annotation Task}

It appears that a major limiting factor to higher annotation-based
learning is the accuracy of the annotators themselves relative to the
evaluation gold standard (the Treebank in this case). To study this
factor, at the end of their active-learning experiments annotators
were asked to annotate a further 100 sentences from the same test data
used to evaluate the learning algorithms.  Their F-measure performance
on this data, as if they were a competing annotation system, is given
in Table \ref{table:annotator_performance}. These measures of
agreement with the gold standard effectively constitutes an upper bound
on the performance of any system trained on their data.

\begin{figure}[htb]
\begin{center}
\mbox{\epsfig{file=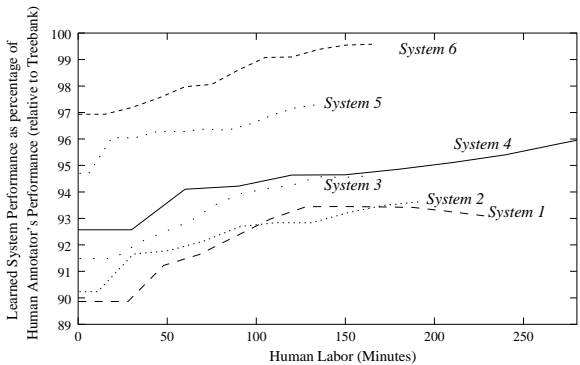, width=3in}}
\caption{\label{figure:annotation-percent}Performance of Ramshaw \& 
Marcus' system trained on annotations by Annotator$_i$, as a
percentage of that person's own performance measured against the
Treebank (an effective upper bound).} 
\end{center}
\end{figure}

Thus to further put annotation-trained system performance in
perspective, Figure \ref{figure:annotation-percent} shows the
performance of individual trained systems relative to the highest
achieved performance of the annotator on which that system was
trained. In each case, the ratio is close to 1, indicating that the
machine learning model achieves performance close to that of the
annotator whose data it was trained on.

%\begin{table}
%\begin{tabular}{|p{3in}||p{3in}|}
%\hline
%Chancellor$_{NNP}$ of$_{IN}$ ( the$_{DT}$ Exchequer$_{NNP}$ )
%Nigel$_{NNP}$ Lawson$_{NNP}$ \ldots &
%\ldots ( vice$_{NN}$ chairman$_{NN}$ ) of$_{IN}$ ( the$_{DT}$
%ad$_{NN}$ agency$_{NN}$ ) ( FCB/Leber$_{NNP}$ Katz$_{NNP}$
%Partners$_{NNPS}$ ) \ldots \\ 
%\hline
%\ldots ( an$_{DT}$ aide$_{NN}$ ) to$_{TO}$ ( Prime$_{NNP}$
%Minister$_{NNP}$ ) ( Shamir$_{NNP}$ ) \ldots &
%( Massachusetts$_{NNP}$ Attorney$_{NNP}$ General$_{NNP}$ James$_{NNP}$
%Shannon$_{NNP}$ ) ,$_,$ \ldots \\
%\hline
%\ldots the$_{DT}$ New$_{NNP}$ York$_{NNP}$ Institute$_{NNP}$ of$_{IN}$
%( Technology$_{NNP}$ ) \ldots &
%\ldots at$_{IN}$ ( Massachusetts$_{NNP}$ Institute$_{NNP}$ ) of$_{IN}$
%( Technology$_{NNP}$ ) .$_.$ \\
%\hline
%Even$_{RB}$ ( the$_{DT}$ president$_{NN}$ 's$_{POS}$ ) doctor$_{NN}$
%\ldots &  
%( Even$_{RB}$ the$_{DT}$ courts$_{NNS}$ ) are$_{VBP}$
%beginning$_{VBG}$ \ldots \\ 
%\hline
%( Tom$_{NNP}$ Wolfe$_{NNP}$ 's$_{POS}$ ) novel$_{NN}$ \ldots &
%( Guillermo$_{NNP}$ Ortiz$_{NNP}$ ) ( 's$_{POS}$ Sept.$_{NNP}$
%15$_{CD}$ Americas$_{NNP}$ column$_{NN}$ ) \ldots \\
%\hline
%\end{tabular}
%\caption{\label{table:treebank_inconsistencies}Sample of Arbitrary
%and/or Inconsistent Annotations in the Treebank}
%\end{table}

\begin{table}
\begin{center}
\begin{tabular}{|c|c|}
\hline
 & F-Measure Performance on \\
 & 100 held-out sentences \\
\hline
Annotator 1 & 92.92 \\ % charles
\hline
Annotator 2 & 92.54 \\ % michele
\hline
Annotator 3 & 91.27 \\ % jeremy
\hline
Annotator 4 & 90.20 \\ % hans
\hline
Annotator 5 & 88.17 \\ % kelly
\hline
Annotator 6 & 86.14 \\ % mihai
\hline 
Annotator 7 & 83.86 \\ % gideon
\hline
\end{tabular}
\caption{\label{table:annotator_performance}Annotation Performance on
100 Test Set Sentences}
\end{center}
\end{table}

\section{\label{section:cost_models}Cost Models for Cross-Modal
Learning Comparison} 

Traditionally, evaluation models in the machine learning literature
measure performance relative to variable quantities of training data.
However, this measure is inappropriate for contrasting data-trained
with rule-writing approaches where time and labor cost are the primary
variables.  Here we present cost models allowing these different
approaches to be compared directly and objectively.

Below are two ways to evaluate the relative cost of a learning algorithm.  
The first measure contrasts performance relative to a common standard
of invested human labor. The second measure considers a fuller range
of potential development costs in a monetary-based common denominator.
%of potential development costs in a monetary-based relative cost measure.

\subsection{Time-Based Cost}

For the purposes of the above experiments, we have chosen total human
effort (in time) as the variable resource, which successfully maps
machine-learning and rule-based learning on a common measure of
training resource investment.

Time-based evaluation is also useful for comparing systems trained on
different annotators.  As shown in Figure \ref{figure:annotation-ruleWriting},
annotators varied in the time they used to tag the full 500-sentence
data set. Because performance tended to be lower when trained
on the faster annotators (perhaps due to less careful work), a 
performance-per-sentences-tagged measure would tend to rank these
individuals lower than their slower (but more careful) colleagues.
However, when measured by the learning accuracy achieved per 
amount of annotator time invested, the faster but noisier annotators
performed much more competitively (although the relative benefits of higher volumes
of noisier data may not extrapolate well). Annotator-time-based performance measures
provide a useful standard for evaluating this volume-noise tradeoff.

\subsection{Monetary Cost}

\begin{table*}
\hspace*{-0.1in}\begin{tabular}{|r@{ =  }l|c|c|}
\hline
\multicolumn{2}{|c|}{Cost Model Parameter} & Annotation &
Rule-writing \\   
\hline
$IDC_{M}$ & Infrastructure Development Cost (for tagging/RW
environment) & Shared  &  Shared \\
$S_O$     &  Number of initial gold standard sentences for
training  & 100 & 100\\
$AC_{TB}$ & gold standard (Treebank) Annotation Cost (per
sentence) & $x$ & $x$ \\
$LC_M$    & Labor Cost for Annotation or RW (per hour) &
\$12.00/hour &\$12.00/hour \\
$MC_{A}$  & Cost of Machine Cycles for Annotation/RW Support &
\$0.24/hour  & \$0.12/hour\\
$T$ & Variable time investment & & \\
\hline
\end{tabular}
\caption{\label{table:money_cost} Example Monetary-Based Cost
Parameters for Model Comparison}
\end{table*}

Annotator labor doesn't capture the complete relative cost of a given approach, however.
It is useful, therefore, to measure resource investment in terms of the
common denominator of monetary cost over a fuller set of potential cost variables.
Table \ref{table:money_cost}
details a set of other monetary parameters considered in the current studies.  Given these
parameters, one possible approximation of this cost function given
variable time investment $T$ and learning method $M$ is:
\begin{eqnarray*}
MonetaryCost(M,T) = & IDC_M + (S_0 + AC_{TB}) \\
& + (T * (LC_M + MC_A))
\end{eqnarray*}

Although we assume equal labor cost rates $LC_M$ for annotation and
rule writing, these may substantially differ in some environments, and
certainly will be higher for professional-quality annotation or
rule-based development. And while the estimates of the machine cycle
cost necessary to support this work on Linux-based PC's vary somewhat,
they are relatively dwarfed by the labor costs. We have assumed that
the infrastructure development costs for the tagging and rule-writing
environments, while initially variable across methods, have already
been borne and to the extent that both interface systems port to new
languages and domains with relative ease, the incremental development
costs for new trials are likely to be relatively low and
comparable. Finally, this cost model takes into account the cost of
developing or acquiring the $S_0$ gold standard tagged data (e.g. from
the Treebank) to provide initial and/or incremental training feedback
to the annotator or rule writer to help force consistency with the
gold standard. We have found that both learning modes can benefit from
this high quality feedback. However, the cost $x$ of developing such a
high-quality resource for new languages or domains is unknown, and
likely will be higher than the non-expert labor costs employed here.

\section{Rules vs. Annotation-based Learning \,---\, Advantages and
Disadvantages} 

In the previous sections, we investigated the performance differences
and resource costs involved for using humans to write rules vs.  using
them for annotations.  In this section, we will further compare these
system development paradigms.

Annotation-based human participation has a number of significant
practical advantages relative to developing a system by manual
rule-writing:

\begin{itemize}

\item Annotation-based learning can continue indefinitely, over weeks and
months, with relatively self-contained annotation decisions at each
point.  In contrast, rule-writers must remain cognizant of potential
previous rule interdependencies when adding or revising rules,
ultimately bounding continued rule-system growth by cognitive load
factors.
%Rule-writers in
%these experiments all stopped development at the point where they felt 
%that adding additional rules ceased to yield performance gain.

\item Annotation-based learning can more effectively combine the efforts
of multiple individuals.  The tagged sentences from different data
sets can be simply concatenated to form a larger data set with broader
coverage.  In contrast, it is much more difficult, if not impossible,
for a rule writer to resume where another one left off.  Furthermore,
combining rule lists is very difficult because of the tight and
complex interaction between successive rules.  Combination of rule
writing systems is therefore limited to voting or similar classifier
techniques which can be applied to annotation systems as well.

\item Rule-based learning requires a larger skill set, including
not only the linguistic knowledge needed for annotation, but also
competence in regular expressions and an ability to grasp the complex
interactions within a rule list.  These added skill requirements
naturally shrink the pool of viable participants and increases their
likely cost.

\item Based on empirical observation, the performance of rule writers
tend to exhibit considerably more variance, while systems trained on
annotation tend to yield much more consistent results.

\item Finally, the current performance of annotation-based training is
only a lower bound based on the performance of current learning
algorithms.  Since annotated data can be used by other current or
future machine learning techniques, subsequent algorithmic
improvements may yield performance improvements without any change in
the data. In contrast, the performance achieved by a set of rules is
effectively final without additional human revision.

\end{itemize}

The potential disadvantages of annotation-based system development for
applications such as base NP chunking are limited.  Given the cost
models presented in Section \ref{section:cost_models}, one potential
negative scenario would be an environment where the machine cost
significantly outweighed human labor costs, or where access to active
learning and annotation infrastructure was unavailable or costly. %Yet
%under foreseeable situations where machine analysis of text is even
%pursued, and assuming public domain access to our annotation and
%active learning toolkits, such a scenario is unlikely.
However, under normal circumstances where machine analysis of text is
pursued, and public domain access to our annotation and active
learning toolkits is assumed, such a scenario is unlikely.

\section{Conclusion}

This paper has illustrated that there are potentially compelling
practical and performance advantages to pursuing active-learning based
annotation rather than rule-writing to develop base noun phrase
chunkers.  The relative balance depends ultimately on one's cost
model, but given the goal of minimizing total human labor cost, it
appears to be consistently more efficient and effective to invest
these human resources in system-development via annotation rather than
rule writing.

\section{Acknowledgements}

The authors would like to thank Jan Hajic, Eric Brill, Radu Florian,
and various members of the Natural Language Processing Lab at Johns
Hopkins for their valuable feedback regarding this work.

\bibliographystyle{acl}

\bibliography{references}

\end{document}